\documentclass[conf]{new-aiaa}
\usepackage[utf8]{inputenc}

\usepackage{graphicx}
\usepackage{amsmath}
\usepackage{siunitx}

\newcommand{\vehicle}{air taxi\xspace}
\newcommand{\vehicles}{air taxis\xspace}
\newcommand{\lidar}{LiDAR\xspace}

\usepackage{multicol}
\usepackage[caption=false,font=footnotesize]{subfig}
\usepackage{upgreek}
\usepackage{todonotes}

\usepackage{ragged2e}
\usepackage{booktabs}
\usepackage[linesnumbered, ruled, vlined]{algorithm2e}
\SetAlgoNlRelativeSize{-5}

\usepackage{xspace}
\newcommand{\eg}{{\it e.g.},\xspace}

\newcommand{\ie}{{\it i.e.},\xspace}
\newcommand{\etc}{{\it etc.~}\xspace}
\newcommand{\ci}{{\it (i) }}
\newcommand{\cii}{{\it (ii) }}
\newcommand{\ciii}{{\it (iii) }}

\title{Synergistic Perception and Control Simplex for \\ Verifiable Safe Vertical Landing}

\author{Ayoosh Bansal\footnote{Graduate Student, Computer Science, 201 North Goodwin Avenue, Urbana, IL 61801, USA.}}
\author{Yang Zhao\footnote{Graduate Student, Mechanical Science and Engineering, 1206 W. Green St. MC 244, Urbana, IL 61801, USA.}}
\author{James Zhu\footnote{Undergraduate Student, Electrical and Computer Engineering, 306 N Wright St, Urbana, IL 61801, USA.}}
\author{Sheng Cheng\footnote{Postdoc, Mechanical Science and Engineering, 1206 W. Green St. MC 244, Urbana, IL 61801, USA.}}
\author{Yuliang Gu\footnote{Graduate Student, Mechanical Science and Engineering, 1206 W. Green St. MC 244, Urbana, IL 61801, USA.}}
\affil{University of Illinois Urbana-Champaign, Urbana, Illinois, 61801}
\author{Hyung-Jin Yoon\footnote{Assistant Professor, Mechanical Engineering, 115 W. 10$^{th}$ St, Cookeville, TN 38505, USA.}}
\affil{Tennessee Tech University, Cookeville, Tennessee, 38505}
\author{Hunmin Kim\footnote{Assistant Professor, Electrical and Computer Engineering, 1501 Mercer University Drive, Macon, GA 31207, USA.}}
\affil{Mercer University, Macon, Georgia, 31207}
\author{Naira Hovakimyan\footnote{W. Grafton and Lillian B. Wilkins Professor, Mechanical Science and Engineering, 1206 W. Green St. MC 244, Urbana, IL 61801, USA.}}
\author{Lui Sha\footnote{Donald B. Gillies Chair in Computer Science, Computer Science, 201 North Goodwin Avenue, Urbana, IL 61801, USA.}}
\affil{University of Illinois Urbana-Champaign, Urbana, Illinois, 61801}%

\begin{document}

\maketitle

\begin{abstract}

Perception, Planning, and Control form the essential components of autonomy in advanced air mobility.
This work advances the holistic integration of these components to enhance the performance and robustness of the complete cyber-physical system.
We adapt \textit{Perception Simplex}, a system for verifiable collision avoidance amidst obstacle detection faults,
    to the vertical landing maneuver for autonomous air mobility vehicles.
We improve upon this system by replacing static assumptions of control capabilities with \textit{dynamic confirmation},
    \ie real-time confirmation of control limitations of the system,
    ensuring reliable fulfillment of safety maneuvers and overrides,
    without dependence on overly pessimistic assumptions.
Parameters defining control system capabilities and limitations, \eg maximum deceleration, are continuously tracked within the system and used to make safety-critical decisions.
We apply these techniques to propose a verifiable collision avoidance solution for autonomous aerial mobility vehicles operating in cluttered and potentially unsafe environments.

\end{abstract}

\section{Introduction}

\lettrine{A}{utonomous} vertical takeoff and landing (VTOL) aerial mobility vehicles, referred to as \vehicles, are poised to revolutionize urban transportation, providing efficient and convenient mobility solutions.
With the removal of human pilots, the autonomy system bears primary responsibility for the safety of the passengers, vehicle and others in its environments.
A safety-critical maneuver with unique challenges is vertical landing an \vehicle.
The ability to land safely in cluttered environments while simultaneously minimizing the landing time is an important milestone toward the viability of air taxis as a mode of mass transportation.
The \vehicle must reliably detect obstacles and ensure any collision avoidance actions are reliably executed by low-level actuator control.
Further, while ensuring collision avoidance and passenger comfort, the descent velocity needs to be maximized, to minimize the landing time.

In modern autonomous systems, the verifiable fulfillment of safety requirements has become increasingly complex due to the reliance on inherently unverifiable deep neural networks (DNNs) for various safety-critical tasks.
DNNs enable capabilities that may have remained unattainable otherwise, \eg scene comprehension and decision-making.
However, the inherent limitations against complete verification and analysis make DNN unsuitable for safety-critical tasks~\cite{heaven2019deep,huang2020survey}.
The use of DNN in safety-critical tasks leads to unpredictable and unreliable behaviors, which is especially impactful during safety-critical maneuvers like landing.

To break this impasse, we adapt solutions from autonomous ground vehicles~\cite{bansal2022verifiable,bansal2022synergistic,mao2023sl1}, which share similar concerns, to provide verifiable fail-safe for obstacle detection and collision avoidance.
A verifiable obstacle detection algorithm is used to detect faults in the DNN-based object detectors.
The faults can then be responded to by a high-assurance model-based feedback controller.
This verifiable capability is then integrated within the autonomous control pipeline for fault detection and correction while retaining the benefits of the higher capabilities offered by DNN-based algorithms.

\begin{figure}[t]
  \centering
  \includegraphics[width=.8\linewidth]{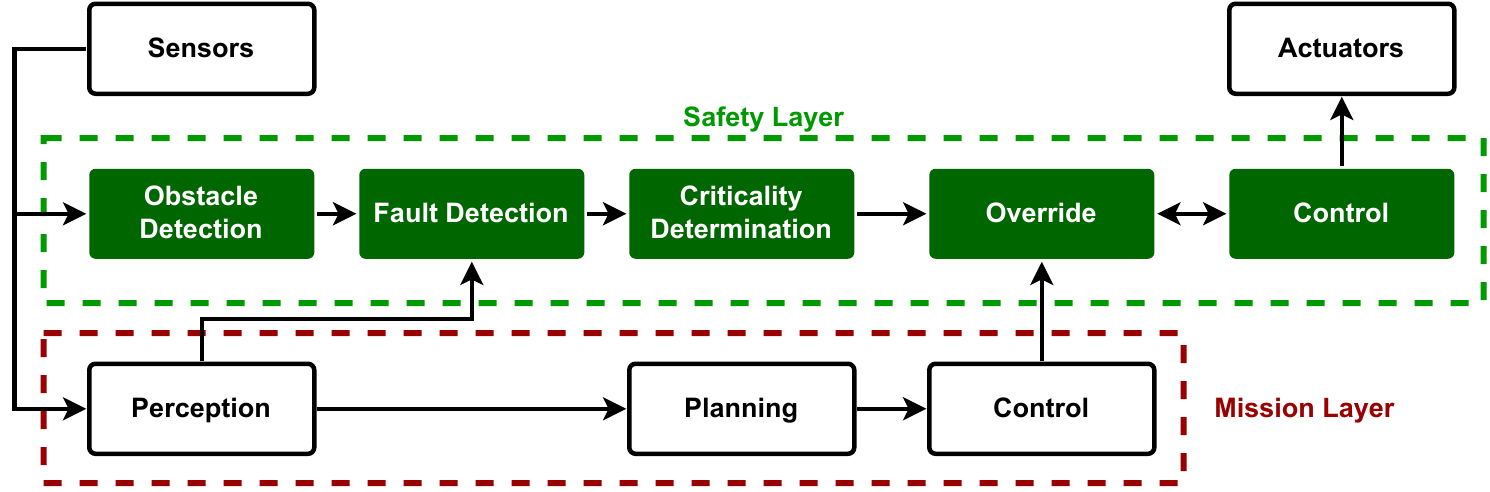}
  \caption{\label{fig:air_taxi_system}
      Proposed system design for autonomous air taxis.
      The mission layer represents the complex, high-performance, although unverifiable, autonomy software.
      The safety layer provides verifiable collision avoidance.
  }
\end{figure}

In addition to adapting these prior solutions for use in \vehicles, this work improves upon them by enabling synergistic cooperation between the different components of the fail-safe mechanism.
A limitation of the prior solutions is the static assumptions about the control capabilities of the system.
We modify this to dynamic confirmations from the control component to others.
Based on the changing control model and environmental conditions the control component broadcasts the changes in key capabilities, \eg max safe acceleration.
We showcase these improvements for the vertical landing scenario.
The proposed system design is illustrated in Figure~\ref{fig:air_taxi_system}.

The key \textbf{contributions} of this work are:

\begin{itemize}
    \item A verifiable solution for safely landing a vertical takeoff and landing urban air mobility vehicles (\vehicles) in cluttered environments, with robust collision avoidance guarantees.
    \item Intercomponent interactions that enhance the system's performance, significantly reducing the landing time, while maintaining high reliability of safety guarantees.
\end{itemize}

\section{Related Work}
\label{sec:related}

The autonomous operation of \vehicles is an active area of research.
Solutions for the safe operation of \vehicles in cluttered environments focus on perception and planning.
Planning a safe path, in cluttered environments, often with limited information is a complex problem,
  with significant prior solutions~\cite{mcintosh2022transition,liao2017motion,moon2018hybrid}.
However, if obstacles are not detected, path planning based on this faulty input cannot ensure collision safety.

Perception for unmanned aerial vehicles has also been studied~\cite{sandino2020uav,patrona2020visual,sonkar2022real}.
While these solutions work well in various contexts, including high-performant mission layers in \vehicles,
the limitations of DNN verification~\cite{heaven2019deep,huang2020survey} make such solutions unreliable for safety critical obstacle detection.
Specifically, landing VTOL vehicles in cluttered environments has been considered in prior work~\cite{rosa2014optical}, however,
the problem and limitations of verifiable perception were not addressed.
For safety-critical obstacle detection, verifiable solutions with precise guarantees must be used.

A recent work on verifiable obstacle detection established a detectability model for a classical obstacle detection algorithm~\cite{bansal2022verifiable}.
This capability is then used to propose Perception Simplex, a verifiable collision avoidance system for ground vehicles~\cite{perception-simplex}.
In Perception Simplex, the control capabilities of the vehicle are assumed to be constant.
Replacing these static conservative assumptions with
  dynamic confirmation via synergistic interactions with low-level control module,
  reducing the pessimism of the system,
  is the key improvement presented in our work.
Furthermore, while the original Perception Simplex was developed and evaluated for ground vehicles only,
  we adapt and apply it to vertical landing maneuver in \vehicle.
The improvements in our dynamic confirmation approach compared to the prior Perception Simplex with static assumptions are presented throughout this work.

The system design proposed in this work also integrates the Control Simplex architecture~\cite{simplex_original,wang2013l1simplex,mao2020finite}, where faults in a complex \textit{high-performance} but unreliable controller are monitored and mitigated by a simpler \textit{high-assurance} controller. Given that control simplex is well established for aerial vehicles, we do not specifically implement or evaluate it.

\section{System Model}

The Simplex design, originally proposed for control systems~\cite{simplex_early,simplex_original}
    and recently for perception faults~\cite{perception-simplex},
    protects an unreliable high-performance mission-critical system using a high-reliability safety-critical layer.
In this work, we present a design for perception and control simplex for \vehicles.
Minimal assumptions are made about the vehicle to support the broad applicability of the proposed solutions.
This work considers a generic \vehicle, capable of vertical take-off and landing (VTOL), similar to prior work~\cite{johnson2018concept,pascioni2022acoustic,nsa_rvtl}, equipped with fully autonomous control.
The vehicle operates in high-density environments, like residences and workplaces.
Therefore, for any maneuver, including take-off and landing, reliable real-time obstacle detection and avoidance are critical requirements for maintaining the safety of the vehicle and any agents in its environment.
Figure~\ref{fig:air_taxi_system} shows an overview of the proposed system design.
We now describe the components of this system, focusing primarily on the safety layer.

\subsection{Mission Layer}
\label{sec:model_mission}

The mission layer hosts high-performance, though potentially low-reliability components.
The mission layer bears full responsibility for the mission of the vehicle, while also trying to maintain safety.
If the mission layer is leading the system to a hazardous state, the safety layer overrides its control outputs, maintaining safety requirements, though the mission of the vehicle may be abandoned.
Complete autonomy for air taxis is an ongoing area of research~\cite{lee2012autonomous,ferrao2022security,chen2022autonomous}.
While developing high-performant mission layer components is an important challenge on its own,
    the mission layer is not the focus of this work.
Therefore, we do not implement or evaluate the mission layer in this work.

\subsection{Sensors}
\label{sec:model_sensor}

\begin{figure}[t]
    \centering
    \includegraphics[width=.5\linewidth]{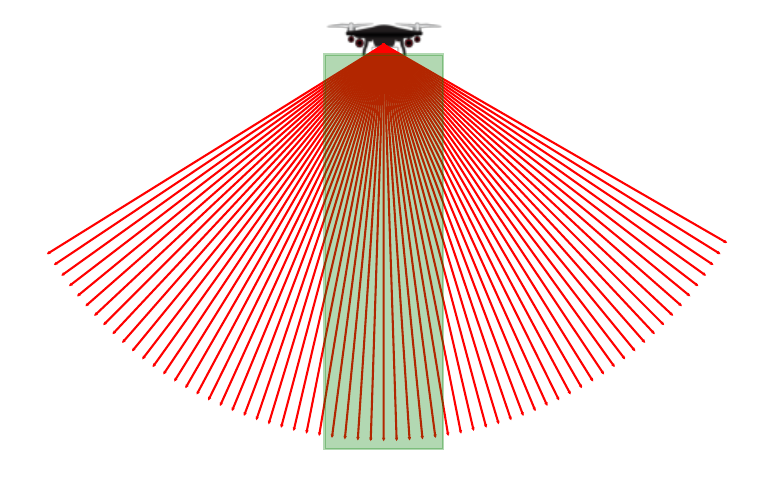}
    \caption{\label{fig:lidar_location}
        \lidar sensor on the \vehicle, marking the vertical landing path as a green area.
    }
 \end{figure}

We assume that the vehicle is outfitted with a \lidar sensor, at the bottom of the \vehicle, as shown in Figure~\ref{fig:lidar_location}.
The potential for use of \lidar in \vehicles has recently been identified~\cite{lidar_article}.
Either rotational or solid-state \lidar can be used to provide the field-of-view as shown in Figure~\ref{fig:lidar_location}.
Note that a scanning sensor (\eg LOAM~\cite{zhang2014loam} or LOWAS~\cite{ramasamy2016lidar}) is unsuitable due to the differences in the use case.
The proximity to obstacles and small margins for maneuvers during landing in cluttered environments necessitates the higher frequency \lidar sensor.

Each laser sensor on the \lidar is given an index $r \in \{0, 1, \cdots N - 1\}$, where $N$ is the number of lasers.
The laser inclination angle set is represented by $\xi$, with angles $\xi_r$ from the horizontal, positive below the horizontal.
The lowest laser below the horizontal is assigned index $r = 0$ and has an inclination of $\xi_0 = 90^o$.
The sensor completes one rotation in $100~ms$ and takes 360 samples, \ie the rotation step angle $\Uppsi = 1^o$.
Rotation steps are indexed with $c$.
Every rotation the sensor returns a range image, a 2D depth image of range values $R_{r,c}~\forall~r\in \{0, 1, \cdots N-1\},~c\in \{0, 1, \cdots \frac{360^o}{\Uppsi}-1\}$.

The \vehicle is also assumed to be capable of sensing its position and velocity. This can be achieved by a combination of sensors \eg inertial measurement unit (IMU), compass, barometer, and global navigation satellite system (GNSS), for measuring and estimating ego state relevant to the inertial frame.

\subsection{Obstacle Detection}
\label{sec:model_vod}

A key requirement for the safety layer is that all software used be certifiable for safety,
    implying that complete logical analysis, verification, and validation must be possible~\cite{feiler2013four,heimdahl2016software}.
Deep learning solutions for obstacle detection do not meet this requirement yet~\cite{heaven2019deep,huang2020survey,pereira2020challenges,willers2020safety,tambon2022certify}.
Therefore, for obstacle detection in the safety layer,
    we use a \lidar-based classical algorithm to detect obstacles~\cite{bogoslavskyi16iros,bogoslavskyi17pfg}.
Briefly, discontinuities in the \lidar returns caused by the existence of obstacles are expressed as geometric relationships and thresholded to distinguish between obstacles and ground.

\begin{align}
    \alpha_{r,c} &= \left\{\begin{array}{lr}
        0^o,\text{if } r=0 \\%
        atan2(|R_{r-1,c} sin \xi_{r-1} - R_{r,c} sin \xi_{r}|, |R_{r-1,c} cos \xi_{r-1} - R_{r,c} cos \xi_{r}|),\text{otherwise}%
        \end{array}\right\}. \label{eq:alpha} \\
    \Delta \alpha_{r,c} &= \left\{\begin{array}{lr}
        0^o,\text{if } r=0 \\%
        |\alpha_{r,c} - \alpha_{r-1,c}|,\text{otherwise}%
        \end{array}\right\}. \label{eq:delta_alpha}
\end{align}

\noindent It is assumed that $r=0$ is on ground, and any return $R_{r,c}$ is from an obstacle if

\begin{align}
    \Delta \alpha_{r,c} &>  \alpha_{th}, \label{eq:alpha_condition}
\end{align}

\noindent where $\alpha_{th}$ is a configurable threshold parameter.

For vertical landing, once the mission layer decides the landing target, it is assumed that $R_{0,c}$ fall near the center of the landing target,
    \ie the \vehicle is positioned vertically above the landing target.
However, if that is not true, the analysis can be started from the return closest to the landing target center.
It is worth noting that obstacles may occupy the landing target, \ie $r=0$ returns may be from an obstacle, and not the landing target.
Also when the \vehicle is at a high enough altitude that the landing target is out of the \lidar range, there is no landing target return, \ie $R_{0,c}$ returns are invalid.
Therefore, the obstacle detection considers three distinct cases:

\begin{itemize}
    \item[\ci]   When the center of the landing target is clear and in range of the \lidar, \ie $R_{0,c}$ returns are valid and from landing target, the algorithm described above applies, \ie \eqref{eq:alpha}, \eqref{eq:delta_alpha}, and \eqref{eq:alpha_condition}~\cite{bogoslavskyi16iros,bogoslavskyi17pfg}.

    \item[\cii]  The above must be disambiguated from when $R_{0,c}$ returns are from an obstacle. This is achieved by comparing the \lidar return range $R_{0,c}$ to the expected distance to the landing target based on the landing target position and the \vehicle position. If this difference is greater than a threshold ($H_{th}$) the landing target is not clear. An Obstacle is on the landing target if
    \begin{align}
        \text{Vehicle Altitude} - \text{Landing Target Altitude} - R_{0,c} &>  H_{th}. \label{eq:landing_target_obstacle}
    \end{align}
   
    \item[\ciii] When the landing target is outside of the range of the \lidar, \ie $R_{0,c}$ returns are invalid, all the other returns are from obstacles.
\end{itemize}

\subsection{Detectability Model}
\label{sec:model_detectability}

The algorithm described above (\eqref{eq:alpha}, \eqref{eq:delta_alpha}, and \eqref{eq:alpha_condition}) was analyzed, with its detectability limits established in a recent work~\cite{bansal2022verifiable}.
The detectability model provides guarantees for obstacle detection, which are used in this work.
However, the prior work focused on ground vehicles only, therefore, we address here the changes necessary to apply the detectability model to \vehicle.
For the detectability limits to be applicable the following constraints must be met:
\begin{itemize}
    \item[\ci]  Within a known range, all invalid \lidar returns must be caused by the laser not encountering any solid surfaces.
    Such range is part of \lidar sensor specifications.
    Within the specification range, returns may still be lost due to weather and visibility effects, however, such effects can be accommodated with reduced effective range, as derived in prior work~\cite[(9)]{perception-simplex}. Adversarial objects with extremely absorbent, dissipating or transparent surfaces could also violate this constraint, though are out of the scope of this work.

    \item[\cii] Obstacle surfaces must be large enough to ensure they do not fit between gaps in \lidar lasers.

    \begin{align}
        \Delta \xi_{r,c} &= |\xi_{r,c} - \xi_{r-1,c}| ~ \forall ~ r \in \{1, 2, \cdots N-1\}. \\
        \text{Each Surface Dimension} &\geq \max(\Uppsi, \max(\Delta\xi_1, \Delta\xi_2, \cdots \Delta\xi_{N-1} ))  R  \frac{\pi}{180^o}, \label{eq:surface_edge}
    \end{align}
    \noindent where $R$ is the range return from the obstacle \ie distance of the obstacle from the sensor.

\end{itemize}

The prior work on verifiable obstacle detection~\cite{bansal2022verifiable} established precise limitations on the minimum size of an obstacle with flat surfaces (\eg cuboid) that is always detected, under specific conditions and constraints, as a function of the distance between the obstacle and the ground vehicle~\cite[Theorem 1, Corollary 1, 2, and 3]{bansal2022verifiable}.
While obstacles may be detected with just one \lidar return from them, obstacles are always detected when two consecutive \lidar points return from the obstacle, and prior noted constraints are met.
Therefore \eqref{eq:surface_edge} can be extended to create a simplified, though pessimistic, detectability model, ensuring that when constraints are met, obstacles are always detected if

\begin{align}
    \text{Each Surface Dimension} &\geq 2 * \max(\Uppsi, \max(\Delta\xi_1, \Delta\xi_2, \cdots \Delta\xi_{N-1} )) R \frac{\pi}{180^o} \label{eq:detectability_2points}
\end{align}

\begin{figure}[t]
    \centering
    \begin{minipage}[t]{0.58\textwidth}
        \centering
        \includegraphics[width=.9\linewidth]{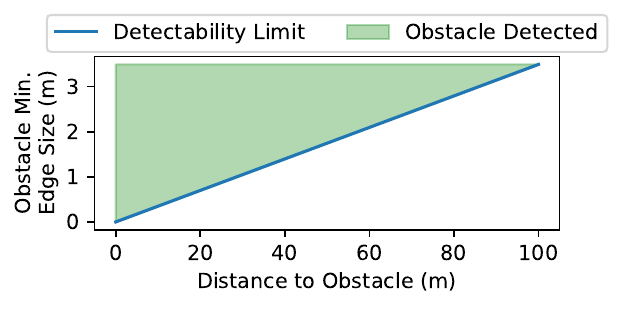}
        \caption{\label{fig:detectability}
          Obstacle size limit for detectability guarantee. The green shaded area signifies the conditions where the obstacle is guaranteed to be detected, given other constraints are also met ($\S$\ref{sec:model_detectability}). Note that this is a pessimistic model, \ie obstacles may still be detected in some parts of the unshaded area of the plot.
        }
    \end{minipage}
    \hfill
    \begin{minipage}[t]{0.4\textwidth}
        \centering
        \includegraphics[width=.9\linewidth]{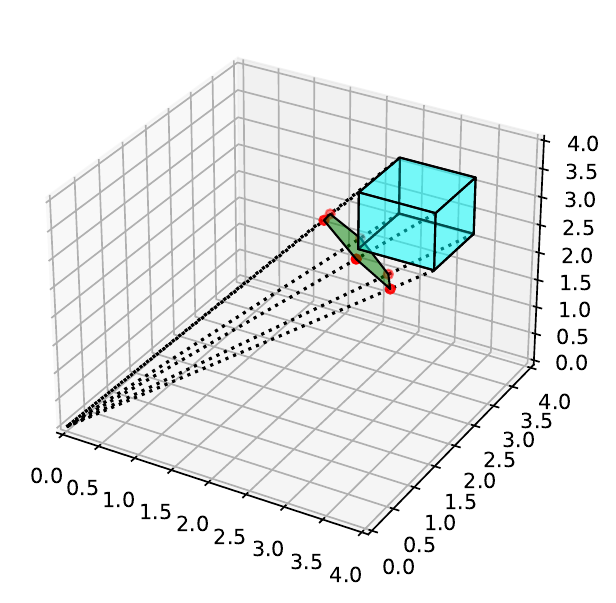}
        \caption{\label{fig:projection}
            Projection of the obstacle to a plane towards the \vehicle (origin), at the obstacle's point closest to the \vehicle.
        }
    \end{minipage}
\end{figure}

Figure~\ref{fig:detectability} plots the detectability limit for the \lidar parameters described in Section~\ref{sec:model_sensor}.
Given a safety policy defining the minimum dimensions of obstacles that must be detected by \vehicles during descent ($\mathcal{E}$), a maximum distance at which obstacles are always detected can be determined as

\begin{align}
    D^{Det} &= \frac{180 \mathcal{E}}{2 \pi * \max(\Uppsi, \max(\Delta\xi_1, \Delta\xi_2, \cdots \Delta\xi_{N-1} ))}.\label{eq:ddet} \\
    H_{th} &= \mathcal{E}.
\end{align}

\subsection{Fault Detection}
\label{sec:model_det_compare}

Faults are determined by comparing the mission layer object detection output against the safety layer obstacle detection results.
To compare these detections a minimized set of requirements, defined in prior works~\cite{bansal2022verifiable,perception-simplex}, are used.
Briefly, rather than 3D bounding boxes, a projection towards the vehicle is considered and compared, as shown in Figure~\ref{fig:projection}.
It is noteworthy that prior works, only consider a line projection for ground vehicles/obstacles due to the lack of motion in the height dimension relative to the ground,
    \vehicles would need to consider a plane, as shown in Figure~\ref{fig:projection}. 
When obstacles detected by the safety layer, are not adequately covered by mission layer detections, 
    a situation can arise where the mission layer plans a path that collides with this obstacle.
Therefore, in such a condition, an obstacle existence detection fault is determined.
However, before an override is initiated, the risk posed by this fault must be considered.

\subsection{Fault Criticality}
\label{sec:model_risk}

The criticality of an obstacle existence detection fault in the mission layer is determined based on the position of the obstacle as detected by the safety layer.
If obstacles are in the landing path of the \vehicle, as shown in Figure~\ref{fig:lidar_location}, the vehicle always responds to such obstacles.
Outside of the landing path, obstacles are evaluated with a physical model for risk, developed in prior work~\cite{bansal2021risk}, which considers all possible positions, \ie existence region~\cite{schmidt2006research}, of the obstacle at different times in the future, checking for collisions with the planned trajectory of the \vehicle.

\subsection{Override}
\label{sec:model_decision}

\begin{algorithm}[t]
    \DontPrintSemicolon
    \SetKwFunction{Match}{isDetected}
    \SetKwFunction{Risk}{isCollisionRisk}
    \SetKwFunction{Control}{isControlHazard}
    \caption{\label{alg:fault_alg}
        Override Decision Logic.
    }

    \KwResult{No Override, Control Override, or Hover}

    \KwIn{
      $O_m$ : Mission layer's current set of objects,~
      $O_s$ : Safety layer's current set of obstacles
    }

    \BlankLine

    \If(\hskip 8mm \hfill \tcp*[h]{Run when either detection set changes.}){$O_m.update()~||~O_s.update()$}
    {
      \ForEach{$o_s$ in $O_s$}{
        \If{\Match{$o_s, O_m$}}{
          continue
        }
        \If{\Risk{$o_s$}}{
          \KwRet Hover \hfill \tcp*[h]{Override trajectory with stop and hover.}
        }
      }
    }
    \If{\Control{}}{
      \KwRet Control Override \hfill \tcp*[h]{Use mission layer trajectory, but safety layer control.}
    }
    \KwRet No Override \hfill \tcp*[h]{Allow mission layer to control actuators.}

    \BlankLine

    \SetKwProg{Fn}{Function}{\string:}{}
    \Fn{\Match{$o_s, O_m$}}
    {
      Safety-critical minimized obstacle detection requirements ($\S$\ref{sec:model_det_compare}).
    }

    \SetKwProg{Fn}{Function}{\string:}{}
    \Fn{\Risk{$o_s$}}
    {
        Potential for Collision using landing trajectory, and Existence Regions~\cite{schmidt2006research,bansal2021risk} ($\S$\ref{sec:model_risk}).
    }

    \SetKwProg{Fn}{Function}{\string:}{}
    \Fn{\Control{}}
    {
      Vehicle approaching control hazard~\cite{wang2013l1simplex,mao2020finite}.
    }    
\end{algorithm}

The override module performs three functions,
    \ci   override the trajectory from the mission layer, to halt and hover when an obstacle detection fault posing a collision risk is detected;
    \cii  switch the low-level control from high-performance mission layer control module to high-reliability safety layer control when the system is at risk of exiting safe control state space;
    \ciii coordinate with the control module to maintain parameters for control capabilities.

While the first two functions have been established in prior works~\cite{perception-simplex,simplex_original,wang2013l1simplex},
    the coordination with control to dynamically confirm capabilities in real-time is the improvement proposed in this work.
The dynamic confirmation is described in detail in Section~\ref{sec:dynamic_confirmation}.
Override module behavior is described in Algorithm~\ref{alg:fault_alg}.

For the stop and hover override to be effective, the detection range from the detectability model ($D^{Det}$) needs to be larger than the stopping distance ($D_{max}^{stop}$).
A maximum safe velocity limit must be imposed,  to keep the current stopping distance below $D_{max}^{stop}$~\cite[Theorem 4]{perception-simplex}.

\begin{align}
    D^{Det} &\geq D_{max}^{stop}. \label{eq:det_constraint}\\
    v^{safe}_{max} &= \sqrt{(a_{max}L_{max})^2+2 a_{max}D_{max}^{stop}} -a_{max}L_{max},  \label{eq:vsafe_clear}
\end{align}

\noindent where $L_{max}$ is the max computation latency of the safety layer, $v^{safe}_{max}$ is the max safe velocity at which collision avoidance can be guaranteed, and $a_{max}$ is the max deceleration which is $a_{max}^{WC}$ with static worst-case assumptions and $a_{max}^{DC}$ when using dynamic confirmation.

\subsection{Control}
\label{sec:model_control}

The control module in the safety layer can receive trajectory commands from the mission layer planner when the system is in the control override state.
The commands may also come from the safety layer override module when, due to perception faults, the system needs to come to a stop and hover.
Independent of the source, once the reference state and control commands are available, 
we can leverage existing multirotor tracking controllers (\eg geometric control~\cite{lee2010geometric}, nonlinear model predictive control~\cite{sun2021comparative}, and incremental nonlinear dynamic inversion~\cite{tal2020accurate}) to track the landing trajectory. However, such tracking controllers are designed for nominal dynamics, which does not account for uncertainties (\eg ground effect, wind) that deteriorate flight performance, even causing catastrophic consequences during landing. Therefore, we use the $\mathcal{L}_1$ adaptive augmentation~\cite{wu20221,wu2023L1QuadFull} to compensate for the uncertainties that the air taxi experiences during landing. The $\mathcal{L}_1$ adaptive control is a robust adaptive control, which is capable of compensating for uncertainties at an arbitrarily high rate of adaptation limited only by the sensors, actuators and computational capabilities. A crucial property of the $\mathcal{L}_1$ control is the decoupling of the estimation loop from the control loop, which allows the use of arbitrarily fast adaptation without sacrificing the robustness of the closed-loop system~\cite{hovakimyan2010L1}. The $\mathcal{L}_1$ adaptive control has been successfully validated on the National Aeronautics and Space Administration's (NASA) AirStar $5.5\%$ subscale generic transport aircraft model~\cite{gregory2009l1,gregory2010flight}, Calspan's Learjet~\cite{ackerman2016l1,ackerman2017evaluation}. Its latest development has been extended to quadrotors~\cite{wu20221,wu2023L1QuadFull}, which significantly improves the tracking performance while providing theoretical guarantees. 

\subsubsection{Dynamics of VTOL vehicle}
The equation of motion of the air vehicle can be derived by Newton's Law. 
We selected the inertial frame with a unit vector $\{i, j, k\}$, pointing to north, east, and up, respectively. 
As we focus on the landing of VTOL vehicles, we assume that there is no rotational movement and horizontal cruise for the vehicle. 
Therefore, we only consider the movement in the vertical direction, and, using Newton's law, the vehicle's equation of motion in the vertical direction of the inertial frame without uncertainties can be written as
\begin{align} \label{EoM_without_uncertainties}
    \dot{p}(t) &= v(t), \\
    \dot{v}(t) &= -g + \dfrac{1}{m}u(t),
\end{align}
where $p(t)$ and $v(t)$ are the position and velocity of the vehicle in the vertical direction, $g = 9.81~m/s^2$ is gravitational acceleration, $m$ is the total mass of the vehicle and $u(t)$ is the force applied on the vehicle in the vertical direction.

\subsubsection{Controller}

\begin{figure}[t]
    \centering
    \includegraphics[width=.95\linewidth]{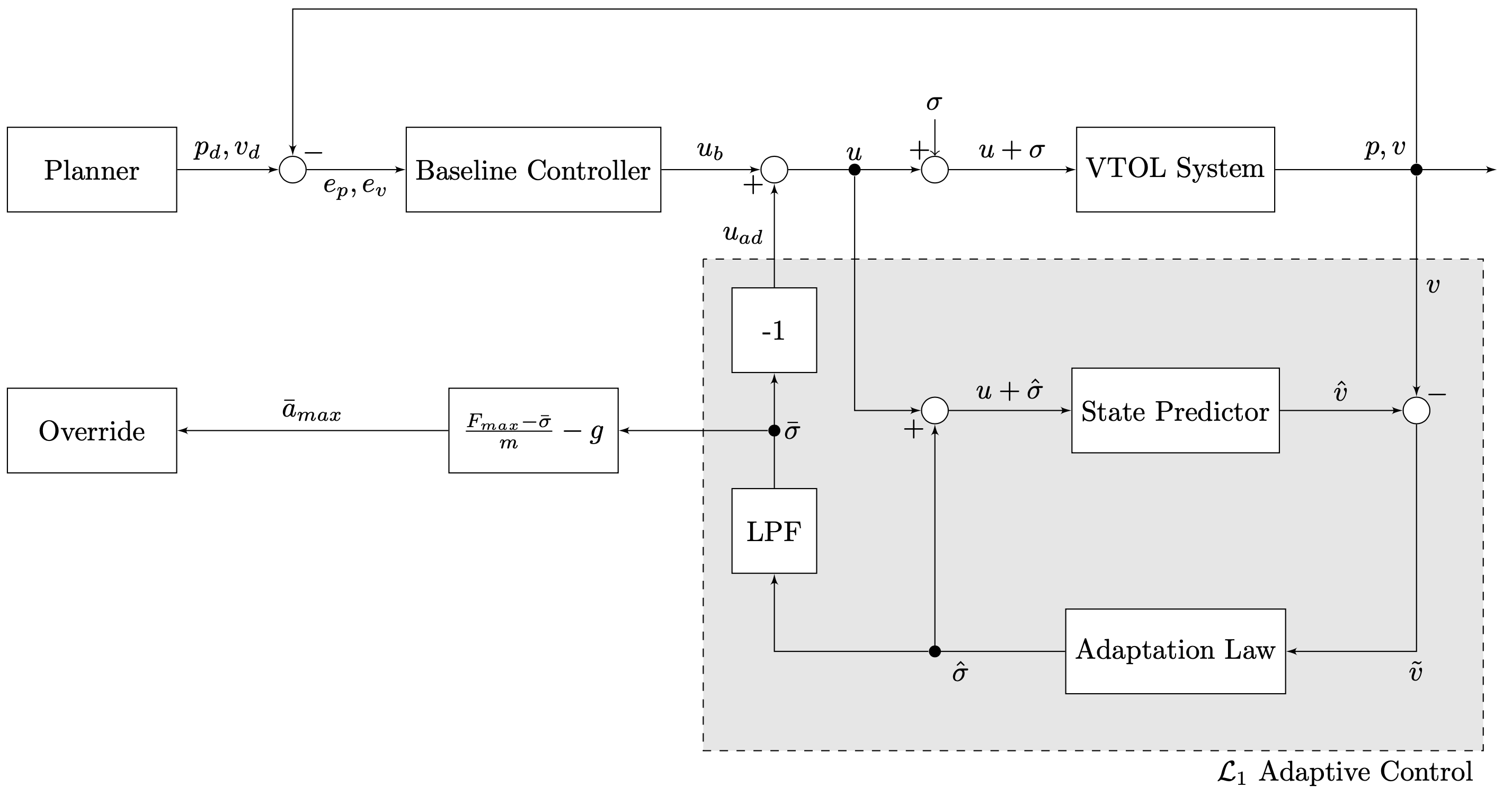}
    \caption{\label{fig:control_diagram}
        The framework of the control module comprises a baseline controller, an $\mathcal{L}_1$ adaptive controller, and the estimation of maximum acceleration $a_{max}$. 
    }
  \end{figure}

The control module comprises primarily of a baseline controller and an $\mathcal{L}_1$ adaptive controller, as shown in Figure~\ref{fig:control_diagram}. While a baseline controller can let the vehicle follow the target position and velocity sent from the planner,  uncertainties, such as wind disturbance, mass deviation, and thruster failure, are much harder to deal with. 
Therefore, in addition to the baseline controller, we use the $\mathcal{L}_1$ adaptive augmentation~\cite{wu2023L1QuadFull} to compensate for uncertainties.
As this work focuses on the landing maneuver, a simplified system dynamics, baseline controller, and $\mathcal{L}_1$ adaptive controller will be presented in this section.
The uncertainties affecting the system can be combined as additional forces in the equation of motion (\ref{EoM_without_uncertainties}).

\begin{align} 
    \dot{p}(t) & = v(t), \label{EoM_with_uncertainties1} \\ 
    \dot{v}(t) & = -g + \dfrac{1}{m} (u_b(t) + u_{ad}(t) + \sigma(t)), \label{EoM_with_uncertainties2}
\end{align}
where the control input is the sum of the baseline controller input $u_b(t)$ and the adaptive controller input $u_{ad}(t)$, and $\sigma(t)$ represents the combined uncertainties that the vehicle experiences.

The baseline control input is calculated with a proportion of the position error and velocity error, in the form of
\begin{align}
    u_b(t) = \begin{bmatrix}
        k_p & k_v
    \end{bmatrix} \begin{bmatrix}
        e_p(t) \\ e_v(t)
    \end{bmatrix},
\end{align}
where $k_p$ and $k_v$ are gains calculated with the LQR method indicated in~\cite{cook2021robust} and~\cite{acheson2021examination}, 
$e_p(t) = p_d(t) - p(t)$ is the position error between the desired position given by the planning layer and the real position, and $e_v(t) = v_d(t) - v(t)$ is the velocity error between the desired velocity and the real velocity.

In addition to the baseline controller, the $\mathcal{L}_1$ adaptive controller uses a state predictor to get predicted velocity $\hat{v}(t)$. 
Comparing the predicted velocity with the real velocity, the $\mathcal{L}_1$ adaptive controller estimates uncertainties with an adaptation law.
Then, based on the estimated uncertainty $\hat{\sigma}(t)$, it generates an adaptive control input $u_{ad}(t)$ to compensate for uncertainties and guarantee the robustness of the vehicle. 
The state predictor gives the expected state of the vehicle with knowledge of dynamics, control inputs $u(t)$, and estimated uncertainties $\hat{\sigma}(t)$:

\begin{align} \label{estimated_v}
    \dot{\hat{v}}(t) = -g + \dfrac{1}{m}(u_b(t) + u_{ad}(t) + \hat{\sigma}(t)) + A_s \tilde{v}(t),
\end{align}
where $\dot{\hat{v}}(t)$ is the derivative of the vehicle's predicted vertical velocity in the inertial frame, $A_s$ is the customized Hurwitz matrix to ensure the stability of the state predictor, and $\tilde{v}(t) = \hat{v}(t) - v(t)$ is the error in the predicted velocity as compared to the real velocity.

Based on the above analysis, the $\mathcal{L}_1$ adaptive controller needs to adjust estimated uncertainty $\hat{\sigma}(t)$ to minimize the state prediction error $\tilde{v}(t)$. 
However, given the discrete nature of software runtime, the controller can only estimate uncertainties in discrete steps. 
To estimate uncertainties in the discrete-time domain, we assume that $T_s$ is the sampling period of the digital system and $n$ is the discrete time index. 
Following the method in~\cite{wu2023L1QuadFull}, in the time period $t \in [nT_s, (n+1)T_s)$, the velocity prediction error $\tilde{v}(t)$ at time $(n+1)T_s$ can be derived by subtracting the real state (\ref{EoM_with_uncertainties2}) to the estimated state (\ref{estimated_v}):
\begin{align} \label{discrete_v_tilde}
    \Tilde{v}((n+1)T_s) & = exp(A_s T_s) \Tilde{v}(n T_s) + A_s^{-1} (exp(A_s T_s) - 1) B\hat{\sigma}(n T_s) - \int_{n T_s}^{(n+1)T_s} exp(A_s ((n+1)T_s - t)) B\sigma(t)dt,
\end{align}
where $B = \dfrac{1}{m}$.
Despite the lack of knowledge on the real uncertainty $\sigma(t)$, we can minimize velocity prediction error $\Tilde{v}((n+1)T_s)$ by canceling first two terms of equation (\ref{discrete_v_tilde}):
\begin{align}
    exp(A_s T_s) \Tilde{v}(n T_s) = - A_s^{-1} (exp(A_s T_s) - 1) B\hat{\sigma}(n T_s).
\end{align}
Finally, by moving $\hat{\sigma}(nT_s)$ to one side, we get
\begin{align}
    \hat{\sigma}(nT_s) = -B^{-1} \Phi^{-1} \mu(nT_s),
\end{align}
where $\Phi = A_s^{-1}(exp(A_s T_s) - 1)$ and $\mu(nT_s) = exp(A_s T_s) \tilde{v}_z(nT_s)$.

Based on the explanation of$\mathcal{L}_1$Quad~\cite{wu2023L1QuadFull}, $\Phi^{-1} exp(A_s T_s)$ is the adaptive gain of the uncertainty estimation. With a chosen $A_s$, a small sampling period $T_s$ may yield a high adaptive gain, which will sequentially generate high-frequency estimated uncertainties, used to reject the real uncertainty. However, the high-frequency control input may reduce the robustness of the system~\cite{wu2023L1QuadFull}. Therefore, it is necessary to implement a low-pass filter to reject the high-frequency estimated signal, such as the sensor noise. The overall adaptive control input in the frequency domain is
\begin{align}
    u_{ad}(s) = -C(s)\hat{\sigma}(s),
\end{align}
where $C(s) = \dfrac{\omega}{s + \omega}$ is the low-pass filter and $\omega$ is the cutoff frequency.

With the baseline controller input $u_b(t)$ and adaptive controller input $u_{ad}(t)$, the propulsion force $u(t) = u_b(t)~+~u_{ad}(t)$ will be applied on the vehicle to follow the designed trajectory and velocity targets.

As this work presents a verifiable solution for collision avoidance during landing, while ensuring safety and efficiency without sacrificing agility, we are primarily interested in the maximum acceleration that the vehicle can reach to successfully perform fast landing or emergency hovering. Assuming the maximum propulsion force of the vehicle's engines is $F_{max}$, the  maximum acceleration is
\begin{align}
    a_{max}(t) = \dfrac{F_{max} + \sigma(t) }{m} - g.
\end{align}

However, by its definition, the real uncertainty $\sigma(t)$, is an unknown.
Therefore, the actual maximum acceleration $a_{max}(t)$ is not calculable, and neither are any derivative values, such as the max safe velocity $v_{max}^{safe}$.
To solve this, we use estimated uncertainty $\hat{\sigma}(t)$ in the $\mathcal{L}_1$ adaptive controller to estimate the maximum acceleration.

\subsubsection{Estimated Maximum Acceleration}
\label{subsec:a_max_DC}
While compensating uncertainties, the $\mathcal{L}_1$ adaptive controller can also estimate reliable maximum acceleration $a_{max}$ of the aerial vehicle with confirmation on dynamics. 
Since the vehicle may encounter various uncertainties, the actual maximum acceleration is different from the ideal value. 
Such uncertainties can be estimated by the adaptive law of the $\mathcal{L}_1$ adaptive controller, denoting by $\bar{\sigma}(t)$, where $\bar{\sigma}(s) = C(s)\hat{\sigma}(s)$ in the frequency domain. The low-pass filter is used to reduce interference of the high-frequency estimations, such as sensor noise.
Hence, the estimated maximum acceleration is
\begin{align} \label{a_max_DC}
  \bar{a}_{max}(t) = \dfrac{F_{max} + \bar{\sigma}(t)}{m} - g.
\end{align}

However, when constrained to static worst-case assumptions, the worst-case maximum acceleration $a_{max}^{WC}$, is calculated by considering the worst case of uncertainties (wind disturbance, thruster failure, \etc) and vehicle mass.
\begin{align}
    a_{max}^{WC} = \dfrac{F_{max} - |d_{max}|}{m_{max}} - g,
\end{align}
where $d_{max}$ is the possible combined disturbances, and $m_{max}$ is the maximum possible mass of the vehicle at full load.

\subsection{Recovery}
\label{sec:model_recovery}

If an override is initiated, the \vehicle comes to a stop and hovers in place, however, it must eventually recover from this condition and land.
By construction, the system recovers if faults subside, giving the control of the \vehicle back to the mission layer.
For example, if the mission layer is able to detect the fault-causing obstacle, or the obstacle itself moves away, the override is lifted and the \vehicle continues normally.
Additionally, if the fault-causing obstacle does not overlap with the \vehicle's landing trajectory, the \vehicle would be able to finish the landing maneuver, albeit with a low descent speed and intermittent acceleration/deceleration.
However, if a fault is persistent and in the planned path of the \vehicle,
    the presented design would need manual intervention to land.
This can be in the form of remote human support.
The key value of the safety layer here is to avoid the collision, reach a safe hovering state, and only then wait for manual intervention.

\section{Dynamic Confirmation}
\label{sec:safety_envelope}
\label{sec:dynamic_confirmation}

The improvement presented in this work is the replacement of static assumptions about the control capabilities of the system with dynamic confirmation,
specifically, the capability of the controller to decelerate the \vehicle with high magnitude, in case of emergencies.
Such measures are required to override the landing maneuver, come to a stop and hover in place, to avoid collisions.
Perception Simplex~\cite{perception-simplex} assumes a constant deceleration limit.
For this system to be robust, an overly pessimistic worst-case value needs to be assumed, considering all potential situations that may arise.
This in turn results in an overly slow descent, as in \eqref{eq:vsafe_clear}.
Therefore, this work proposes dynamic confirmation between control and override modules, to replace pessimistic worst-case assumptions for deceleration with 
observed worst-case.

\subsection{Sliding Window}
\label{sec:window}

\begin{figure}[t]
  \centering
  \includegraphics[width=.4\linewidth]{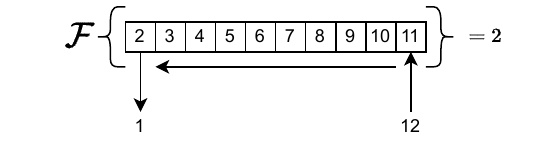}
  \caption{\label{fig:window}
      Sliding window for control parameters which maintains a configurable number of samples of most recent values.
      As a new value is inserted, the oldest one is removed.
      The window returns ($\mathcal{F}$) the worst-case value within it,
        \eg the minimum value for deceleration that the low-level control can provide.
  }
\end{figure}

The override module maintains a history of the control capability parameters received, \eg $\bar{a}_{max}(t)$.
This history is maintained as a sliding window, as shown in Figure~\ref{fig:window}.
At any instance, the worst-case value from the window is chosen to calculate any derived limits or decisions, \eg \eqref{eq:vsafe_clear}.
This simple structure provides several benefits:
\begin{itemize}
  \item[\ci]   If the system or environmental state degrades, causing new observed worst-case values, the sliding window returns ($\mathcal{F}$) the new observed worst-case values immediately.
  \item[\cii]  The sliding window size ($\mathcal{W}$) allows for a limited time continuation of prior observed worst-case values, leading to a delayed recovery from disturbances. This is especially suitable for temporally proximate repetitive disturbances.
  \item[\ciii] The window size ($\mathcal{W}$) can be adjusted dynamically to modify the history duration to be considered.
\end{itemize}

\subsection{Safety Envelope}

\begin{figure}[t!]
  \subfloat[$a = a_{max}^{WC}, L = 0$\label{subfig-se:ps_c}]{%
    \includegraphics[width=0.45\textwidth]{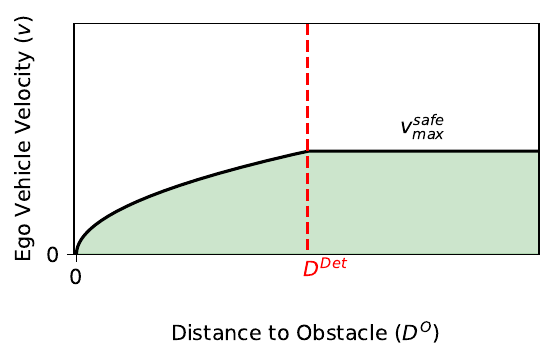}
  }
  \hfill
  \subfloat[$a = a_{max}^{DC}, L = 0$\label{subfig-se:ss_c}]{%
    \includegraphics[width=0.45\textwidth]{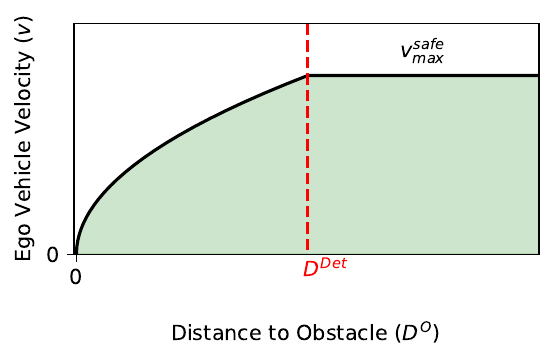}
  }
  \\ 
  \subfloat[$a = a_{max}^{WC}, L = L_{max}$\label{subfig-se:ps_l}]{%
    \includegraphics[width=0.45\textwidth]{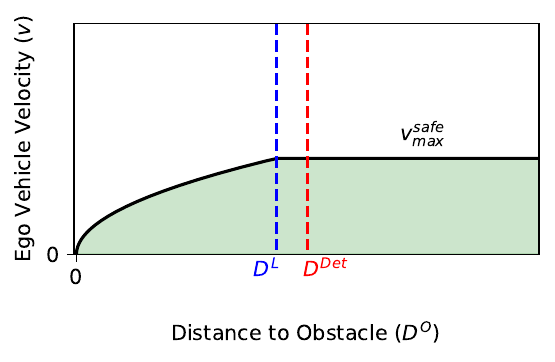}
  }
  \hfill
  \subfloat[$a = a_{max}^{DC}, L = L_{max}$\label{subfig-se:ss_l}]{%
    \includegraphics[width=0.45\textwidth]{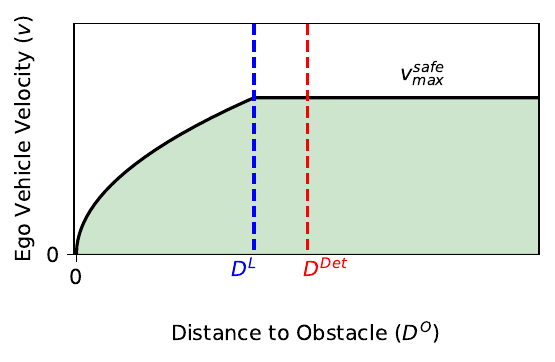}
  }
  \caption{
    These figures represent the safe operating region or safety envelope
    for Perception Simplex for different max decelerations ($a_{max}$) and compute latency ($L$).
    $a_{max} = a_{max}^{WC}$ signifies static worst-case assumptions, while $a_{max} = a_{max}^{DC}$ shows the effect of dynamic confirmation.
    Each plot (solid line) shows the max safe velocity and a green region showing the
    safe operating space where collisions can be avoided by Perception Simplex.
    $D^{Det}$ marks the detectability distance \ie the maximum distance at which obstacles are always detected.
    When compute latency is not considered the safety constraint is $D^{Stop} \leq D^{Det}$.
    However, compute latency between the time the obstacle is first detectable and when the decision based on such detection is available,
    reduces the reaction distance available to the vehicle, changing the constraint to $D^{Stop} \leq D^{L} < D^{Det}$.
    In this illustration, $a_{max}^{DC}/a_{max}^{WC} = 3$, a ratio approximately equal to the ratio of these values in evaluation ($\S$\ref{sec:eval}).
  }
  \label{fig:safety_envelope}
\end{figure}

We now describe the safety envelope of the Perception Simplex system, with static assumptions and the proposed dynamic confirmation.
A safety envelope represents the safe operating state of the vehicle.
When the vehicle state is within this envelope, the simplex system's safety guarantees are valid.
Figure~\ref{fig:safety_envelope} illustrates the safety envelopes for various Perception Simplex setups for the \vehicle.
Note that the safety envelope discussion here is limited to Perception Simplex and Dynamic Confirmation, as these are the focus of this work.
The discussion of safety envelope for control simplex is omitted here as it is well established in prior works~\cite{simplex_early,simplex_original,simplex0,wang2013l1simplex,mao2020finite,mao2023sl1}, and this work does not modify or evaluate it.
To describe the safety envelope, the following parameters must be considered

\noindent%
\begin{tabular}{l c p{.85\linewidth}}
  $D^O$ & : & Distance to an obstacle. \\
  $D^{Det}$ & : & The detection range is based on the detectability model. While this range is assumed a constant here, as prior works discuss, this range may reduce based on weather conditions~\cite{perception-simplex}. \\
  $D^{Stop}$ & : & The current stopping distance of the \vehicle, \ie the distance traveled before coming to a stop, where stopping is defined as hovering in place. \\
  $v$ & : & Descent velocity of the \vehicle. \\
  $v_{max}^{safe}$ & : & Max safe descent velocity at which collision avoidance by stopping and hovering is guaranteed. \\  
  $a_{max}^{WC}$ & : & The max deceleration possible, using worst-case assumptions. \\
  $a_{max}^{DC}$ & : & The observed max deceleration, from the sliding window, \ie $\mathcal{F}(\mathcal{W})$. By definition $|a_{max}^{DC}| \geq |a_{max}^{WC}|$. \\
  $L_{max}$ & : & The maximum computational latency, from sensor input to the resultant actuation commands.
\end{tabular}

\subsubsection{Perception Simplex}

The obvious constraint for avoiding collisions by stopping is that the stopping distance be less than the distance to the obstacle.
Perception Simplex extends this with $D^{Det}$, \ie max distance at which obstacles are guaranteed to be detected.

\begin{align}
  D^{Stop}        & \leq D^O. \label{eq:stop_constraint} \\
  D^{Stop}_{max}  & \leq D^{Det}. \label{eq:det_stop}
\end{align}

Consider the safety envelope shown with a green area in Figure~\ref{subfig-se:ps_c}.
When $D^O \leq D^{Det}$, the obstacle is always detected, by definition of $D^{Det}$.
Therefore, using \eqref{eq:stop_constraint} the velocity constraint for safety can be derived.
However, when $D^O \geq D^{Det}$ the obstacle may not be detected and the vehicle has no method to measure how far away an undetected obstacle is.
Therefore, given the perception system limits, to avoid false negative detection faults, the worst-case of obstacle being present just beyond $D^{Det}$ must be assumed.
Combining this we get

\begin{align}
  v \leq \left\{\begin{array}{lr}
    \sqrt{2*a_{max}^{WC}*D^O},     &\forall D^O \leq D^{Det} \\%
    \sqrt{2*a_{max}^{WC}*D^{Det}}, &\forall D^O > D^{Det}
    \end{array}\right\}. \label{eq:envelope_ps_c}
\end{align}

\subsubsection{Dynamic Confirmation}

The key difference made by dynamic confirmation comes from the improved estimation of $a_{max}$.
Recall that by definition $|a_{max}^{DC}| \geq |a_{max}^{WC}|$.
Figure~\ref{subfig-se:ss_c} shows the change in the safety envelope when $\frac{a_{max}^{DC}}{a_{max}^{WC}} = 3$.
\begin{align}
  v \leq \left\{\begin{array}{lr}
    \sqrt{2*a_{max}^{DC}*D^O},     &\forall D^O \leq D^{Det} \\%
    \sqrt{2*a_{max}^{DC}*D^{Det}}, &\forall D^O > D^{Det}
    \end{array}\right\}. \label{eq:envelope_ss_c}
\end{align}

\subsubsection{Computational Latency}

Another factor that influences the stopping distance of the \vehicle is the computational latency,
specifically, the time duration between the sensor input and the actuation command to stop, when the decision is made based on the data from this sensor input. This latency ($L$) can vary with a large range in the mission layer.
However, the safety layer is a hard-deadline real-time system, therefore $L$ is bounded with a max limit $L_{max}$.
At max safe velocity, the stopping distance changes due to $L_{max}$,

\begin{align}
  D^{Stop} = D^{L} = v_{max}^{safe} * L_{max} + \frac{(v_{max}^{safe})^2}{2a_{max}}. \label{eq:dstop_c_l}
\end{align}

This change in stopping distance and the resultant effect on the safety envelope is illustrated in Figure~\ref{subfig-se:ps_l}~and~\ref{subfig-se:ss_l}.

\subsection{Trade-offs}
\label{sec:unsafe}

\begin{figure}[t]
  \centering
  \includegraphics[width=.6\linewidth]{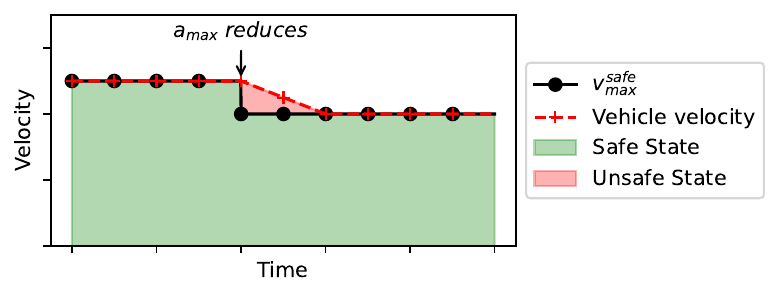}
  \caption{\label{fig:unsafe}
    Transient unsafe state when the system is converging to a reduced $v_{max}^{safe}$, in response to external disturbances.
  }
\end{figure}

While dynamic confirmation improves the performance of the Perception Simplex system, it does create a possibility for unsafe conditions, as shown in Figure~\ref{fig:unsafe}.
Consider a situation where the vehicle descending with $v = v_{max}^{safe}$ experiences sudden large disturbances, causing the max deceleration ($a_{max}$) to reduce significantly.
This change is immediately communicated to the override module, which updates the sliding window, and thus $v_{max}^{safe}$ is also updated accordingly.
However, given that the \vehicle is a physical system with inertia, the vehicle velocity cannot immediately match the reduced $v_{max}^{safe}$.
In this transient state, the safety constraint in \eqref{eq:det_stop} is violated, \ie $D^{Stop} \geq D^{Det}$.

It should be noted that a potential collision in the transient unsafe state requires several precise conditions to occur simultaneously.
While not guaranteed, collisions may be avoided in the transient state if the obstacle is detected early,
  recall that the detectability model is pessimistic (Section~\ref{sec:model_detectability}, Figure~\ref{fig:detectability}), therefore,
  obstacles further away from the detectability limit may still be detected.
Similarly, the control limits are also conservative, further reducing the probability of collisions in the transient state.

This discussion also assumes that $a_{max}^{WC}$ is the true worst-case value, which itself is complicated to ascertain, given different kinds of faults and failures that can happen in \vehicles.
For example, as the physical system degrades over time, the engine thrust reduces.
However, if such possible degradations are considered in the static worst-case formulation, it becomes extremely pessimistic.
An advantage of using dynamic confirmation is the continuous adaptation to the changing control capabilities of the \vehicle, 
  as determined by the adaptive control module, while still operating on a temporally localized observed worst-case.
Furthermore, future works will address such transitory states, as elaborated upon in Section~\ref{sec:future}.
\section{Evaluation}
\label{sec:eval}

\begin{figure}[t]
    \centering
    \includegraphics[width=.7\linewidth]{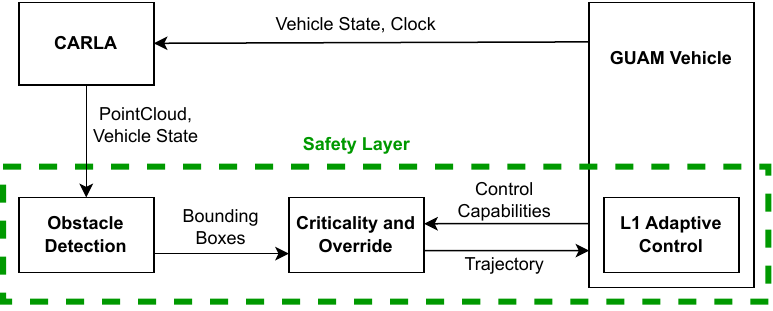}
    \caption{\label{fig:sim}
        Software-in-the-loop simulation setup for evaluation.
    }
\end{figure}

We evaluate the proposed system using a software-in-the-loop simulation,
  running landing maneuvers with $a_{max}$ based on static assumptions and dynamic confirmation.
Different evaluation scenarios change the obstacle position.
Each simulation run is evaluated for safety and performance metrics.
The simulations yield the expected result of no collisions and faster landing when dynamic confirmation is used.

\subsection{Simulator Setup}
\label{sec:eval_setup}
The software-in-the-loop simulation setup, used for this evaluation, is illustrated in Figure~\ref{fig:sim}.
We extend a simulation setup for ground vehicles, developed in a prior work~\cite{simulator1}, to support \vehicles.
CARLA Simulator~\cite{dosovitskiy2017carla} provides the scenario, environment, obstacles and, sensor data.
However, the CARLA simulator was primarily developed for ground vehicles.
Therefore, a high-fidelity MATLAB Simulink Generic Urban Air Mobility (GUAM) vehicle model\footnote{The vehicle model, developed by NASA, is closed-source.} is used here to provide the physical model and control dynamics for the \vehicle.

We integrate CARLA and GUAM vehicle model using Robot Operating System (ROS)~\cite{quigley2009ros}, version Noetic.
The GUAM model computes the vehicle state, which is then replicated within the CARLA environment.
It is noteworthy that the two simulators operate asynchronously, which presents a challenge in achieving reliable integration.
This is mediated by introducing a clock signal from the vehicle model, the average iteration interval of which is used by other components.
However, the variance in the model run-time, coefficient of variation up to $9.29\%$, is still a source of error within the simulation setup.

We integrate $\mathcal{L}_1$ Adaptive Control within the GUAM model to provide the high-reliability control.
In consideration of passenger comfort, to avoid aggressive acceleration, we saturate the maximum propulsion force $F_{max}$ so that the maximum ideal acceleration $F_{max}/m-g$ is limited to $4.69~m/s^2$.
The worst case acceleration $a_{max}^{WC}$ is estimated to be $1.34~m/s^2$ with the assumptions of worst-case operating conditions of wind speed $6~m/s$~\cite{wind_1,wind_2}, additional $10\%$ mass due to full occupancy, and two engines failing, similar to prior work~\cite{beiderman2021hazard}.
For the baseline controller, the control gains $k_p$ and $k_v$~\cite{cook2021robust,acheson2021examination} are integrated within the GUAM model.
For the $\mathcal{L}_1$ adaptive controller, we set $A_s = -50,~T_s = 0.005~s$, and the cutoff frequency of the low-pass filter $C(s)$ as $100~Hz$.
With the above setup, the controller receives position and velocity targets, compares them with the vehicle's current position and velocity, and adjusts propulsion to minimize the state error.

Obstacle detection is adapted from the depth-clustering algorithm~\cite{bogoslavskyi16iros,bogoslavskyi17pfg}.
We also implement the criticality and override modules.
As noted in Section~\ref{sec:model_mission}, authors are unaware of an open-source high-performant mission layer implementation for \vehicles. Additionally, the mission layer is not a focus of this work, therefore, a mission layer is not included in this evaluation setup.
To accommodate for that, any obstacles detected by the safety layer are assumed to be false negative faults in the mission layer.
The causes of such faults in DNN-based object detectors, an active area of research itself~\cite{tu2020physically,miller2022s},
    are out of the scope of this work.
It is assumed that the intended landing target, a mission layer output, is available at the start of the evaluation as is vertically below the current vehicle position.
Finally, we implement a simple landing trajectory planner, to replace the mission layer functionality for trajectory planning.
The planner accelerates the vehicle to $v_{max}^{safe}$, maintains the velocity and as the \vehicle approaches the landing target, reduces the \vehicle velocity to land with a small landing velocity.
The simulations are conducted on a desktop running Ubuntu 20.04 LTS.

\subsection{Metrics}
\label{sec:metrics}

Each simulation run is evaluated for two criteria, safety and performance.
Safety is defined as collision avoidance, \ie throughout the landing maneuver it is checked if any collisions occur.
All simulations yielded the expected result that no collisions were observed, matching the results for ground vehicles from prior work~\cite{perception-simplex}, when $v \leq v_{max}^{safe}$.
The focus of dynamic confirmation is to improve the performance of the safety layer while maintaining the safety guarantees of Perception Simplex.
Performance is defined here as the time duration for the landing maneuver.

\subsection{Scenario: No Obstacle}
\label{sec:eval_clean}

\begin{figure}[t!]
  \subfloat[Deceleration Limit $ a_{max}^{WC} = 1.34 m/s^2$\label{subfig:clean_wc}]{%
    \includegraphics[width=0.48\textwidth]{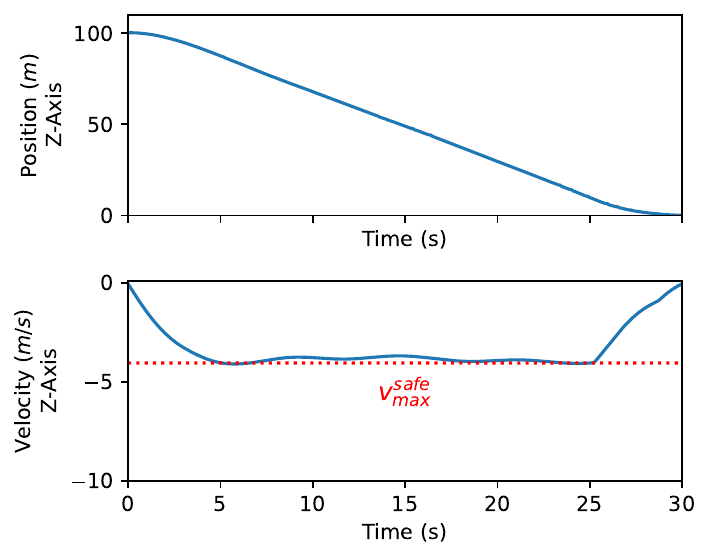}
  }
  \hfill
  \subfloat[Deceleration Limit $  a_{max}^{DC} = 4.25 m/s^2$ (avg.)\label{subfig:clean_dc}]{%
    \includegraphics[width=0.48\textwidth]{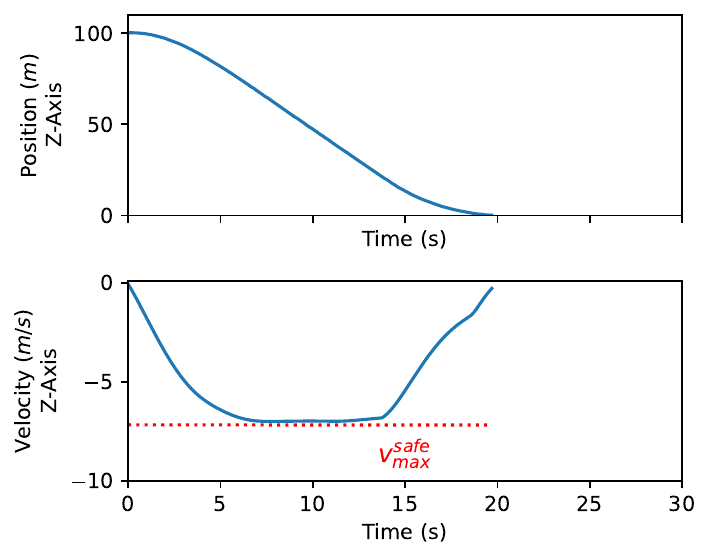}
  }
  \caption{
    Landing maneuver when no obstacles are in the path of the \vehicle.
    With dynamic confirmation, the landing time reduces by $\mathbf{34.2\% (1.52\times)}$.
  }
  \label{fig:eval_clean}
\end{figure}

{\it Experiment.}
The simulation starts with the ego vehicle $100~m$ above the landing site.
We use the coordinate system with the positive z-axis pointing upwards.
The position is measured in a world frame, \ie position is measured from a fixed point on the map, and the ground level is $Z = 0$.
The vehicle starts at X and Y origin, \ie $X = 0,~Y = 0$.
There are no obstacles in the vicinity of the vehicle or the landing path.
Therefore, the start conditions are $v(0) = 0,~p(0) = (0,0,100)$ in $(x, y, z)$ coordinates.

{\it Observations.}
Figure~\ref{fig:eval_clean} shows the position $p$ and velocity $v$ of the \vehicle, along the vertical Z axis.
The vehicle first accelerates till  $v_{max}^{safe}$, then starts decelerating close to the ground, eventually landing with a small landing velocity.
The difference in $v_{max}^{safe}$ limits, originating from the difference in the $a_{max}$, results in the vehicle taking $34.2\% (1.52\times)$ less time for the whole maneuver when using dynamic confirmation (Figure~\ref{subfig:clean_dc}), than when using static worst-case assumptions (Figure~\ref{subfig:clean_wc}).

{\it Discussion.}
The landing maneuver here starts at a relatively low altitude of 100 meters,
  compared to the up to 1.5 km (5000 ft) operating altitude for \vehicles~\cite{ventura2020computational}.
Furthermore, the maneuver starts with 0 initial velocity.
Finally, while the ideal maximum acceleration is $4.69~m/s^2$, environmental disturbances approximating wind speed of $3~m/s$ result in the reduced $a_{max}^{DC} = 4.25~m/s^2$ on average.
Despite these limiting factors, dynamic confirmation speeds up the time to complete this maneuver by $34.2\% (1.52\times)$.

\subsection{Scenario: Obstacle Vertically Below Air Taxi}
\label{sec:eval_below}

\begin{figure}[t!]
  \subfloat[Deceleration Limit $  a_{max}^{WC} = 1.34 m/s^2$\label{subfig:below_wc}]{%
    \includegraphics[width=0.48\textwidth]{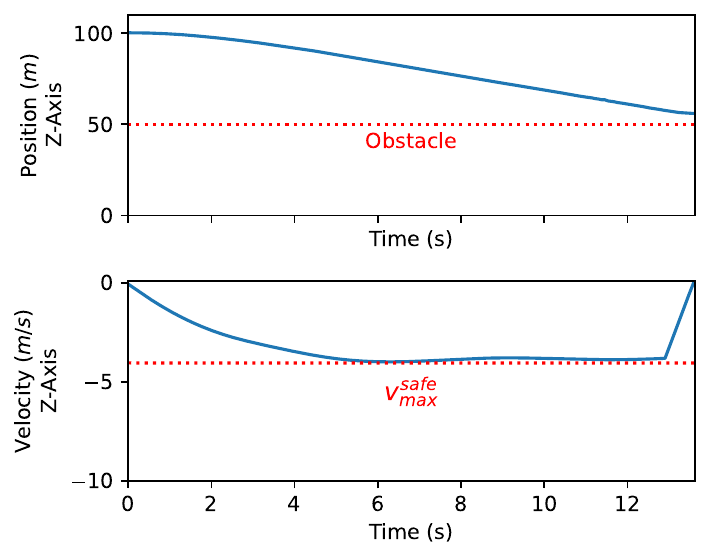}
  }
  \hfill
  \subfloat[Deceleration Limit $ a_{max}^{DC} = 4.25 m/s^2$ (avg.)\label{subfig:below_dc}]{%
    \includegraphics[width=0.48\textwidth]{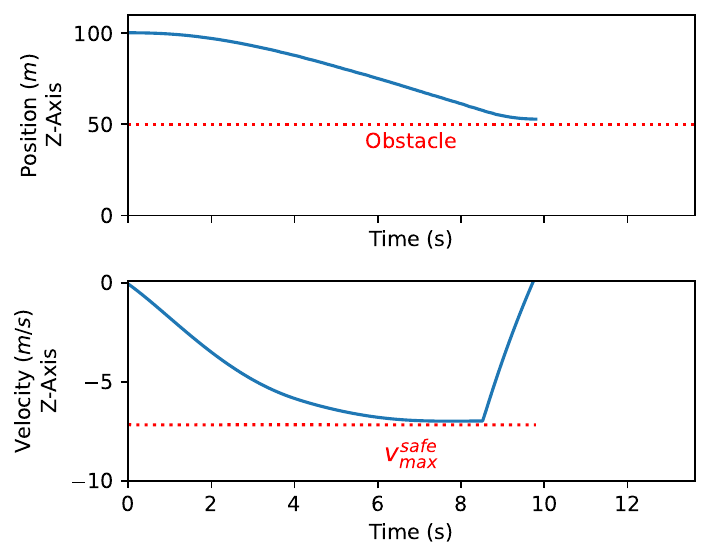}
  }
  \caption{
    Landing maneuver when an obstacle is directly below the \vehicle. 
    Collision is avoided in both cases.
  }
  \label{fig:eval_below}
\end{figure}

{\it Experiment.}
This scenario is the same as the previous one, except an obstacle is introduced vertically below the \vehicle.
The obstacle is at the position of $50~m$ above the ground, and the same horizontal position as the \vehicle,
  \ie obstacle position is $(0, 0, 50)$.

{\it Observations.}
Figure~\ref{fig:eval_below} shows the results.
The obstacle position is marked with a red dotted line.
In both cases, the collision is avoided.
The \vehicle uses the full emergency deceleration to rapidly come to a stop and hover in place.
When operating with static worst-case assumptions, Figure~\ref{subfig:below_wc}, the vehicle comes to a stop with a much higher deceleration than the assumed worst-case.
Therefore, the air taxi comes to a stop with a larger distance to the obstacle (Figure~\ref{subfig:below_wc}),
In comparison, when using dynamic confirmation,
  the observed worst-case is a much closer estimate of the achievable deceleration (Figure~\ref{subfig:below_dc}).

{\it Discussion.}
In this scenario, as the obstacle is vertically below the \vehicle, therefore, it is not detected by the depth clustering algorithm.
As discussed in Section~\ref{sec:model_vod}, this limitation is handled by comparing \lidar returns $R_{0,c}$ with the distance to the landing target, \ie \eqref{eq:landing_target_obstacle}.
While the obstacle is detected well in advance, by design, the safety layer only responds to it as the obstacle distance comes close to the distance to stop.
Therefore, the \vehicle stops just before the obstacle.
This is necessary to allow the mission layer to detect and evade the obstacle, if it can, which could result in the successful completion of the landing.
However, if the mission layer does not detect the obstacle, as the window to avoid the collision becomes smaller, the safety layer initiates the override.
Once the collision has been avoided, external support is required to recover and then land, as discussed in Section~\ref{sec:model_recovery}.

\subsection{Scenario: Obstacle In Path}
\label{sec:eval_inpath}

\begin{figure}[t!]
  \subfloat[Deceleration Limit $  a_{max}^{WC} = 1.34 m/s^2$\label{subfig:inpath_wc}]{%
    \includegraphics[width=0.48\textwidth]{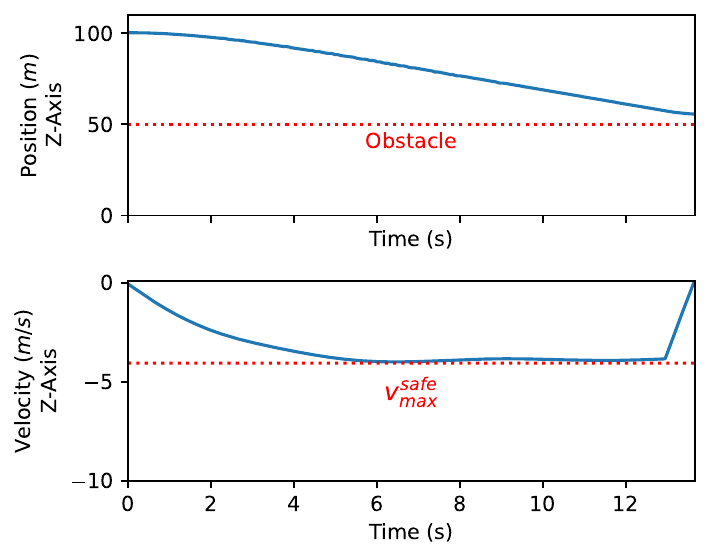}
  }
  \hfill
  \subfloat[Deceleration Limit $ a_{max}^{DC} = 4.25 m/s^2$ (avg.)\label{subfig:inpath_dc}]{%
    \includegraphics[width=0.48\textwidth]{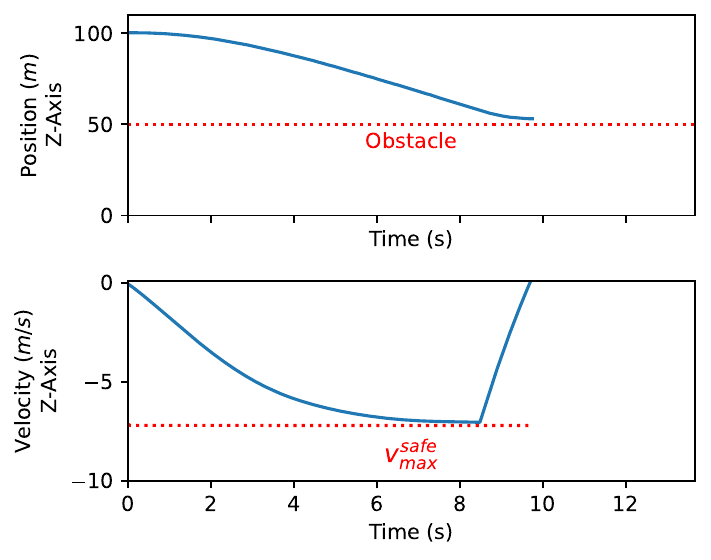}
  }
  \caption{
    Landing maneuver when an obstacle is in the path of the \vehicle. 
    No Collisions were observed.
  }
  \label{fig:eval_inpath}
\end{figure}

{\it Experiment.}
This scenario is the same as the previous one, except that the obstacle is moved forward $2~m$ along the X axis, \ie obstacle position is $(2, 0, 50)$.

{\it Observations.}
Figure~\ref{fig:eval_inpath} shows the result. The \vehicle stops before the obstacle, avoiding a collision.

{\it Discussion.}
The main point of difference from the previous scenario is that here the obstacle bounding box is detected by the depth clustering algorithm~\cite{bogoslavskyi16iros,bogoslavskyi17pfg}, \ie \eqref{eq:alpha}, \eqref{eq:delta_alpha}, and \eqref{eq:alpha_condition}.
By construction of the detectability model and velocity limit $v_{max}^{safe}$, we ensure that the obstacle is detected in time to stop before it and avoid collision.

\subsection{Scenario: Obstacle Not In Path}
\label{sec:eval_notpath}

\begin{figure}[t!]
  \subfloat[Deceleration Limit $  a_{max}^{WC} = 1.34 m/s^2$\label{subfig:notpath_wc}]{%
    \includegraphics[width=0.48\textwidth]{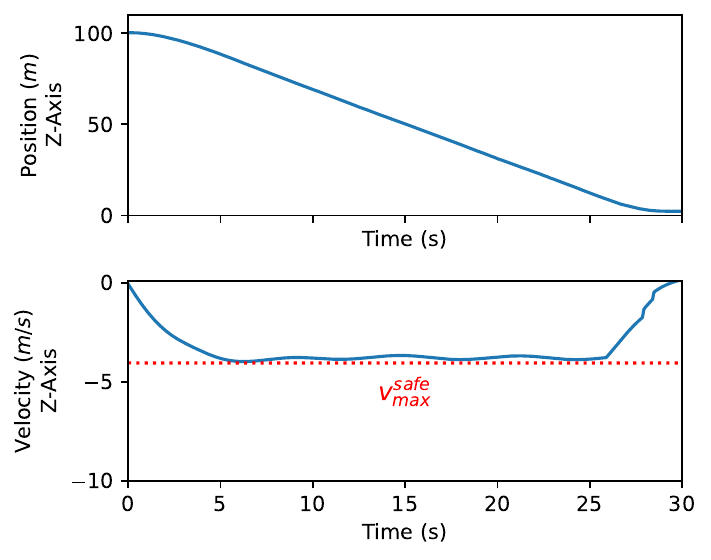}
  }
  \hfill
  \subfloat[Deceleration Limit $ a_{max}^{DC} = 4.25 m/s^2$ (avg.)\label{subfig:notpath_dc}]{%
    \includegraphics[width=0.48\textwidth]{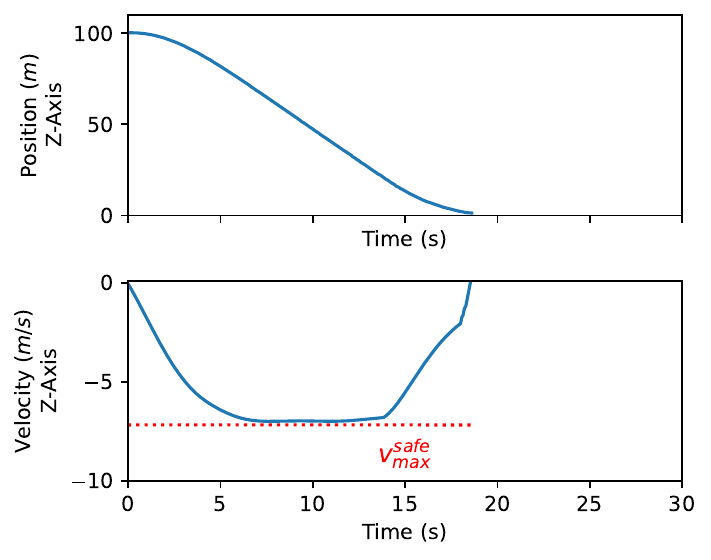}
  }
  \caption{
    Landing maneuver when obstacles are not directly in the path of the \vehicle. 
  }
  \label{fig:eval_notpath}
\end{figure}

{\it Experiment.}
In this scenario, the obstacle is moved 1 meter further ahead, \ie obstacle position is $(3, 0, 50)$.

{\it Observations.}
The obstacle is no longer in the path of the \vehicle. Therefore, the obstacle is detected, and given that it is not in the path, the landing maneuver completes normally.

{\it Discussion.}
The safety layer only overrides the landing maneuver when there is a risk of collision with the obstacle.
Therefore, a stationary obstacle outside of the landing path does not trigger an override,
with the behavior here closely matching when no obstacles were present (Figure~\ref{fig:eval_clean}).

\section{Discussion}
\label{sec:discussion}

\subsection{Verifiability}

One of the main differences between the safety layer design presented in this work and DNN-dependent mission layer systems is verifiability.
All components of the safety layer can be logically analyzed and root causes for any implementation faults can be found precisely.
Therefore, the safety layer implementation can be verified and validated.
This is in contrast to DNN solutions that, as yet, cannot be fully analyzed or verified~\cite{katz2017reluplex,gharib2018safety,liu2021algorithms}.

\subsection{Roles and Responsibilities}

In the presented Perception and Control Simplex design, the safety and mission layers have different roles and responsibilities.
The safety layer is responsible for providing verifiable safety properties, whereas
    the mission layer is responsible for the mission of the vehicle and a best-effort to maintain safety.
When faults in the mission layer risk violating the safety properties guaranteed by the safety layer, it overrides the mission layer, maintaining system safety, but potentially abandoning the mission of the \vehicle.
The safety layer is verifiable fail-safe, best-effort fail-operational.
Therefore, mission layer reliability is still an important factor in making \vehicle a viable autonomous system.

\subsection{Extended Requirements}

This work focuses exclusively on collision avoidance and descent velocity, \ie safety and performance, respectively.
However other requirements may also exist that impact the presented design.
For example, for rider comfort the max acceleration may be limited to a value lower than what is achievable.
This can be addressed by adding a constraint on $a_{max}$, consequently saturating $F_{max}$.
Furthermore, the trajectory planner can be made to avoid the comfort limit, only when required to avoid collisions,
    though given the safety guarantees that would be a rare occurrence.

\subsection{General Applicability}

The chosen application for this work is \vehicles. However, dynamic confirmation can help improve any system Perception Simplex may be used with, \eg autonomous ground vehicles. Changing road surface or brake conditions, that could lead to changes in braking deceleration for the vehicle, could be addressed with dynamic confirmation.
In general, other systems may also benefit from dynamic confirmation, replacing the observed worst-case heuristic for the sliding window used in this work, with other heuristics suitable to the particular system.
Furthermore, the principles of verifiable redundant solutions and synergistic cooperation can be applied to other safety-critical challenges in \vehicles.

\subsection{Static vs Dynamic Acceleration Estimation}
    
The estimation of $a_{max}^{WC}$ is static and can be used as a pessimistic safe constraint.
However, its conservative nature sacrifices the performance of the vehicle. 
In contrast to the static estimation, the involvement of the $\mathcal{L}_1$ adaptive controller can dynamically confirm the system property with consideration of the real-time uncertainties. 
Unexpected behaviors of the vehicle will be detected and combined into $\bar{\sigma}(t)$.
Hence, the dynamic estimation of the $a_{max}$ can maximize the vehicle's performance while ensuring safety, significantly improving efficiency.

\subsection{Impacts}

The max operating velocity $v_{max}^{safe}$ is a constraint required for the validity of safety guarantees of Perception Simplex.
However, the trade-off is that this velocity limit must be maintained, independent of the current state of the system, even in the absence of any current faults.
This work aims to provide a different tradeoff, increasing the $v_{max}^{safe}$ limit,
    while still maintaining the high reliability of Perception Simplex,
    except in extreme combination of circumstances ($\S$\ref{sec:unsafe}).
The increased $v_{max}^{safe}$ implies reduced travel time and improved capacity for any vertiport or landing pad.
In this work, a speed-up of $1.5\times$ was observed ($\S$\ref{sec:eval_clean}),
    despite the starting velocity being 0 and
    the starting position for the landing maneuver being 100 meters,
    compared to the up to 1.5 km (5000 ft) operating altitude for \vehicles~\cite{ventura2020computational}.

\section{Conclusion and Future Work}
\label{sec:conclusion}
\label{sec:future}

This work presents a verifiable solution for collision avoidance during landing maneuver for \vehicles.
It also advances the integration between the safety-critical components, synergistically improving the performance of the system.
The proposed solution is implemented and evaluated using a software-in-the-loop simulation.

There are various directions for further improvements to the proposed system.
Continuing with the holistic integration between the safety layer components,
    ongoing works aim to enable the proactive adaptation of the controller model to upcoming environmental changes,
    with the help of perception to perceive such upcoming changes.
This adaptation will reduce the jitter caused by environmental changes and improve the responsiveness of the control component.
It will further reduce the probability of large unforeseen changes in environmental conditions leading to unsafe conditions ($\S$\ref{sec:unsafe}).
Similarly, while this work considers the combined usage of Perception and Control Simplex,
    the two systems and their override decision and switching remain largely independent.
In future works, we will study closer integration between the two simplex systems to improve upon the presented design.

\section*{Acknowledgments}

This material is based upon work supported by
the National Aeronautics and Space Administration (NASA) under the cooperative agreement 80NSSC20M0229 and University Leadership Initiative grant no. 80NSSC22M0070, and
the National Science Foundation (NSF) under grant no. CNS 1932529 and ECCS 2311085.
Any opinions, findings, conclusions or recommendations expressed in this material
are those of the authors and do not necessarily reflect
the views of the sponsors.%

\bibliography{ref,ref-Sheng, ref-Yang}

\end{document}